\documentclass[11pt]{article}

\usepackage[final]{acl}

\usepackage{times}
\usepackage{latexsym}

\usepackage[T1]{fontenc}

\usepackage[utf8]{inputenc}

\usepackage{microtype}

\usepackage{inconsolata}

\usepackage{graphicx}

\usepackage{hyperref}
\usepackage{url}
\usepackage{amssymb}
\usepackage{algorithm}
\usepackage{algorithmicx}
\usepackage{algpseudocode}
\usepackage{booktabs,multirow,makecell,xcolor,colortbl,siunitx,tabularx}
\usepackage{float}
\usepackage{subcaption}
\usepackage{comment}
\usepackage{tcolorbox}
\usepackage{fontawesome5}

\usepackage{soul}
\usepackage{tikz}
\soulregister\cite{1}
\soulregister\ref{1}
\soulregister\textbf{1}
\soulregister\textit{1}
\soulregister\emph{1}

\title{Improving the Throughput of Diffusion-based Large Language Models \\ via a Training-Free Confidence-Aware Calibration}

\author{
\textbf{Jucheng Shen}\textsuperscript{1,*} \quad
\textbf{Gaurav Sarkar}\textsuperscript{2,\dag} \quad
\textbf{Yeonju Ro}\textsuperscript{3,*} \quad
\textbf{Sharath Nittur Sridhar}\textsuperscript{2} \\
\textbf{Zhangyang Wang}\textsuperscript{3} \quad
\textbf{Aditya Akella}\textsuperscript{3} \quad
\textbf{Souvik Kundu}\textsuperscript{2,*} \\
\textsuperscript{1}Rice University \quad
\textsuperscript{2}Intel \quad
\textsuperscript{3}The University of Texas at Austin \\
\texttt{jucheng.shen@rice.edu}, \texttt{yro@cs.utexas.edu}, \texttt{souvikk.kundu@intel.com}
}

\begin{document}
\maketitle
\renewcommand{\thefootnote}{\fnsymbol{footnote}}
\footnotetext[1]{Corresponding authors.}
\footnotetext[2]{Work done when the author was affiliated with Intel.}
\renewcommand{\thefootnote}{\arabic{footnote}}
\setcounter{footnote}{0}
\begin{abstract}
We present \textbf{CadLLM}, a \textit{training‑free} method to accelerate the inference throughput of diffusion-based LLMs (dLLMs). We first investigate on the dynamic nature of the token unmasking confidence across blocks and steps. Based on this observation, we then present a lightweight adaptive approach that can control the generation \emph{block size}, \textit{step size}, and \emph{threshold} based on the average confidence score of the unmasked tokens. We further reduce the softmaxing overhead of token probability generation by dynamically leveraging a subset of \emph{vocabulary size} to regulate sampling breadth. CadLLM is a plug-and-play model-agnostic with KV caching based dLLMs. Extensive experiments on four popular tasks demonstrate the efficacy of CadLLM to yield throughput improvements ranging from 1.1--2.28$\times$ over the state-of-the-art baseline with competitive accuracy.
\hspace{0.5em}
\href{https://github.com/juchengshen/CadLLM}{\faGithub~Code}
\end{abstract}

\section{Introduction}
\label{sec:intro}

Masked diffusion language models (dLLMs) have advanced rapidly, with open-sourced models such as LLaDA \cite{nie2025largelanguagediffusionmodels} and DREAM \cite{ye2025dream7bdiffusionlarge} demonstrating strong generative capabilities \cite{austin2021d3pm,campbell2022ctmdm}. However, unlike autoregressive language models that generate tokens sequentially in a single forward pass, diffusion-based models rely on a multi-step denoising Markov process that iteratively refines noisy latent states to generate clean text. This stochastic, multi-round refinement incurs substantial computational overhead during inference ~\cite{ye2025dream7bdiffusionlarge, nie2025largelanguagediffusionmodels}, often causing significant slowdown in token generation. 

To improve inference throughput, recently fast-dLLM \cite{wu2025fastdllm} proposed parallel decoding that unmasks all positions above a \emph{static} global confidence\footnote{In this work we define confidence score as the model's probability output score for its predicted tokens at each masked position.} threshold in parallel. While effective, this static thresholding has few key limitations. Firstly, fast-dLLM uses fixed block sizes with fixed step-size per block, essentially ignoring the change in confidence over the sequence. Secondly, it treats each block uniformly with a fixed sampling breadth~\cite{fan2018hierarchical,holtzman2018learning}, ignoring potential certainty differences. Thirdly, it employs a fixed token commit threshold, which also doesn't account for the variation in confidence across inference steps.

\noindent
\textbf{Our Contributions.} To address the above limitations, in this paper we present \textbf{CadLLM}, a \textit{training-free adaptive method} that uses lightweight confidence signals to allocate compute where uncertainty persists and save it where predictions stabilize. Specifically, we first analyze the difference in confidence across different blocks and denoising steps. Interestingly, we find that the confidence varies significantly across these dimensions (refer to 
Section~\S\ref{sec:case}). To more optimally balance the compute for token denoising in dLLM, we then present CadLLM. Based on the average confidence of the unmasked tokens in a block, CadLLM dynamically allocates the \textit{block size}, \textit{step size} and unmasking \textit{threshold} during the inference steps. To further reduce the token prediction softmax operation cost we present an \textit{adaptive vocabulary subset} selection based on the prediction confidence. 

Extensive experiments across multiple datasets at different generation lengths ($g$) demonstrate the efficacy of CadLLM with throughput gains ranging from \textbf{1.1--2.28}$\times$ over the state-of-the-art fast dLLM baseline \cite{wu2025fastdllm} on LLaDA, and \textbf{1.1--1.4}$\times$ on DREAM, with competitive accuracy.



\newcommand{\good}[1]{\textcolor{green!60!black}{#1}}
\newcommand{\bad}[1]{\textcolor{red!70!black}{#1}}
\newcommand{\midq}[1]{\textcolor{orange!70!black}{#1}}
\newcommand{\cmark}{\good{$\checkmark$}}
\newcommand{\xmark}{\bad{$\times$}}
\newcommand{\omark}{\midq{$\bigcirc$}}

\begin{figure*}[!th]
  \centering
  \begin{subfigure}[t]{0.69\linewidth} 
    \centering
    \includegraphics[width=\linewidth]{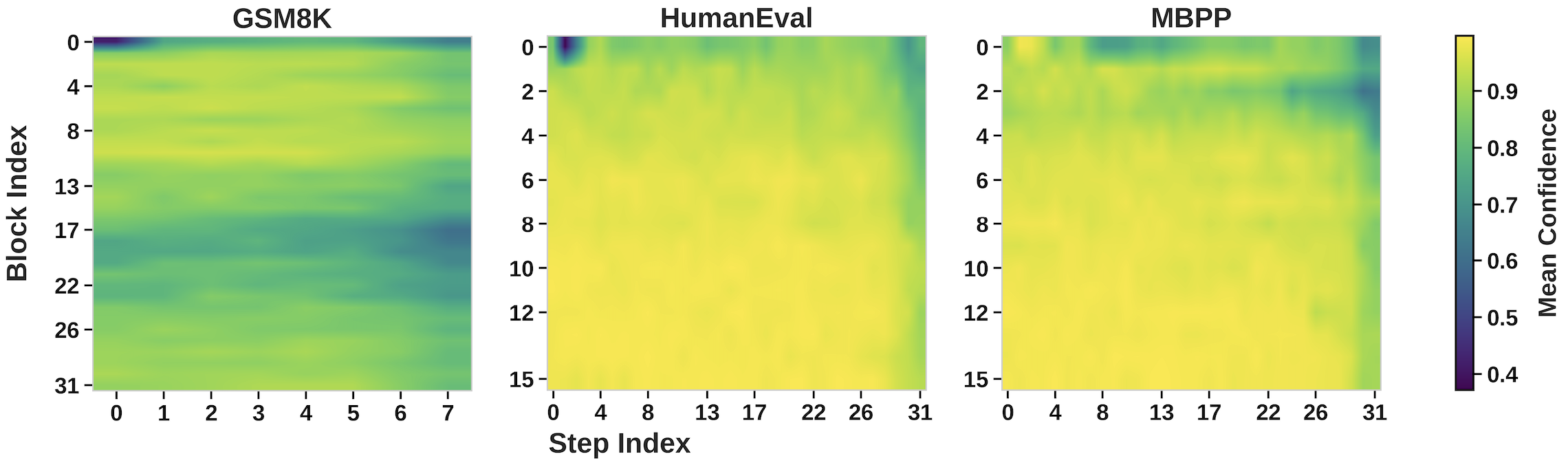}
    \caption{Per-block, per-step confidence heatmaps. Lighter the color the higher the confidence.}
    \label{fig:step_block_heatmap}
  \end{subfigure}
  \hfill
  \begin{subfigure}[t]{0.30\linewidth}
    \centering
    \includegraphics[width=\linewidth]{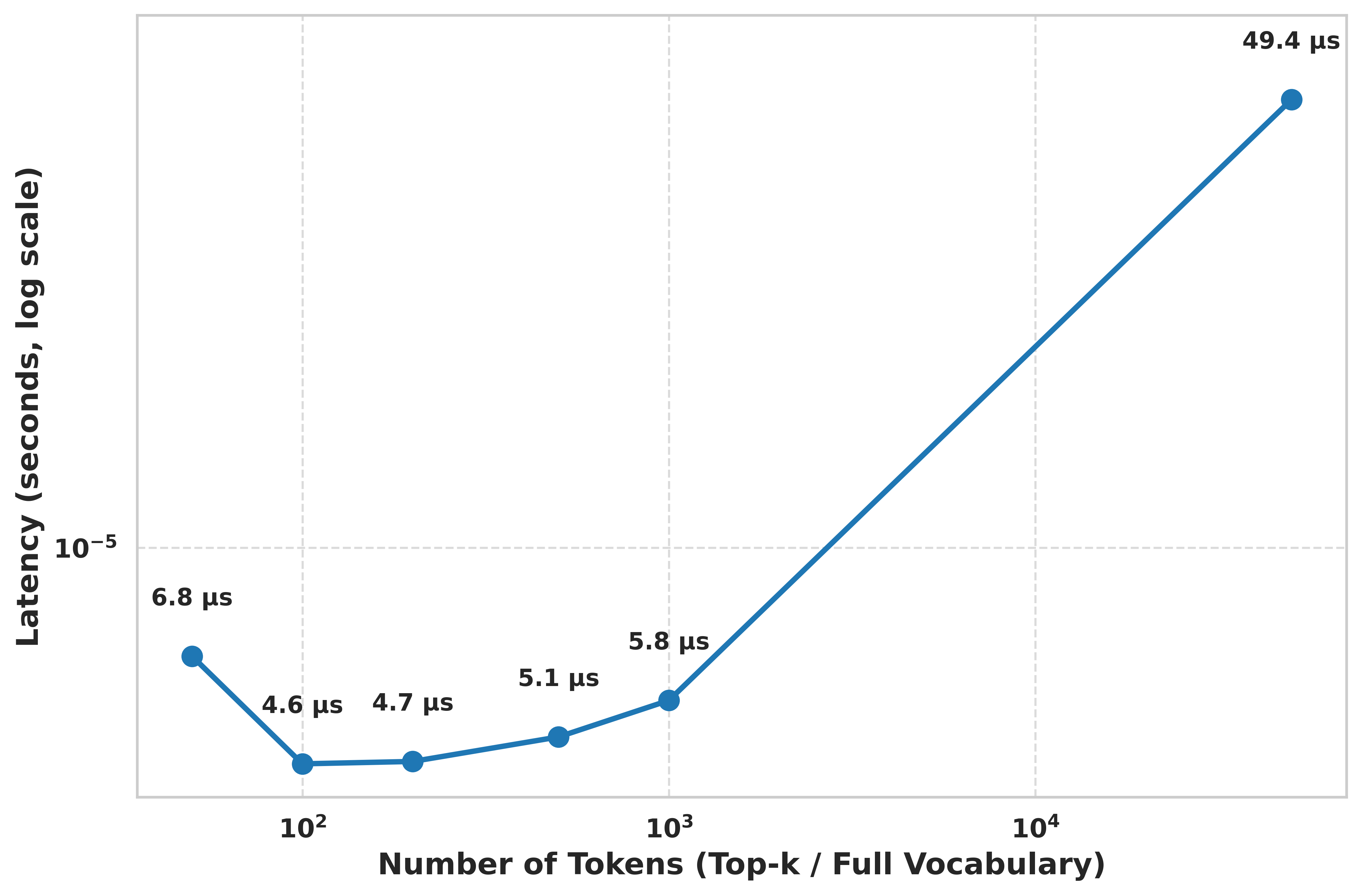}
    \caption{Softmax latency.}
    \label{fig:vocab_size_latency}
  \end{subfigure}
  \vspace{-1em}
  \caption{(a) Confidence dynamics for three difference datasets. (b) Latency vs.\ vocabulary size.}
  \label{fig:heatmap_latency}
  \vspace{-1.5em}
\end{figure*}
\section{Preliminaries}
\label{sec:prelim}
Generation in masked diffusion language model (MDM) is modeled by a network $p_\theta$ that predicts clean tokens for masked positions conditioned on the partially noised context and (optionally) the time $t$ \cite{austin2021d3pm}.  
At each step, given the current evidence $E = (x_t, t)$, the model computes $p_\theta(X^i = v \mid E)$ for each position $i$ and selects tokens with the highest confidence $c^i(E) = \max_{v \in \mathcal{V}} p_\theta(X^i = v \mid E)$ to unmask, iteratively updating the sequence until reaching $t = 0$ or a stopping condition.
Fast-dLLM \cite{wu2025fastdllm} accelerates MDM generation by decoding masked tokens in parallel according to confidence signals. It organizes the output into $K$ blocks of size $B$ and performs at most $S$ refinement steps per block  via MDM mechanism. 

\paragraph{Threshold-Based Rule.}
For a working sequence $x$, it compute confidences $\{c^i(E)\}_{i\in\mathcal{M}(x)}$ and unmask all positions whose confidence exceeds a fixed global threshold $\tau\in(0,1)$ \cite{wu2025fastdllm}:
\begin{equation}
\small
\mathcal{U}_\tau(E)=\{\, i \in \mathcal{M}(x) \,:\, c^i(E)\ge \tau \,\}    
\end{equation}

\paragraph{Factor-Based Rule.}
Factor-based rule controls the degree of parallelism via a \emph{factor} parameter $\phi>0$. For a sorted confidence order $c^{(1)}\ge \cdots \ge c^{(m)}$ for the $m$ masked positions in the current block and choose the largest $r$ satisfying:
\[
(r+1)\left(1-c^{(r)}\right) < \phi,
\]
It then unmasks the top-$r$ positions. Intuitively, this decodes more tokens when individual confidences are uniformly high and fewer when they are borderline. 


\section{Motivational Case Study}
\label{sec:case}
\paragraph{Confidence is Not Uniform.}
Figure~\ref{fig:step_block_heatmap} shows per-block, per-step confidence for three datasets. Within each block, confidence rises quickly then plateaus; across blocks, difficulty varies—some stabilize immediately, others remain uncertain. This relationship between confidence and decoding difficulty is independently corroborated by concurrent work~\cite{lu2026adablockdllmsemanticawarediffusionllm}, which shows that fixed block sizes force the model to commit low-confidence tokens prematurely, leading to higher decoding error. A fixed schedule thus over-refines easy blocks while under-serving harder ones. When confidence is high, large block size can exploit parallel decoding since more tokens exceed the threshold; when low, smaller blocks help stabilize outputs faster. Similarly, the step size should be large under low confidence and small once predictions are stable. Finally, the confidence threshold itself must adapt—should be high at situations to prevent premature commitments, and should be low at situations to  accelerate decoding. 

\paragraph{Sampling Breadth has Latency Overhead.}
Figure~\ref{fig:vocab_size_latency} evaluates the softmax latency for different vocabulary sizes. It shows that the latency grows sharply with vocabulary size: evaluating all $\sim$50K tokens is nearly an order of magnitude slower than using a small subset. Since softmax is called at every step, this may become a major bottleneck. This necessitates the adaptive vocabulary size selection to perform the softmax computation on only a subset of total vocabulary. 



\section{CadLLM Methodology}
\label{sec:method}

\begin{figure}[!t]
  \centering
  \includegraphics[width=\linewidth]{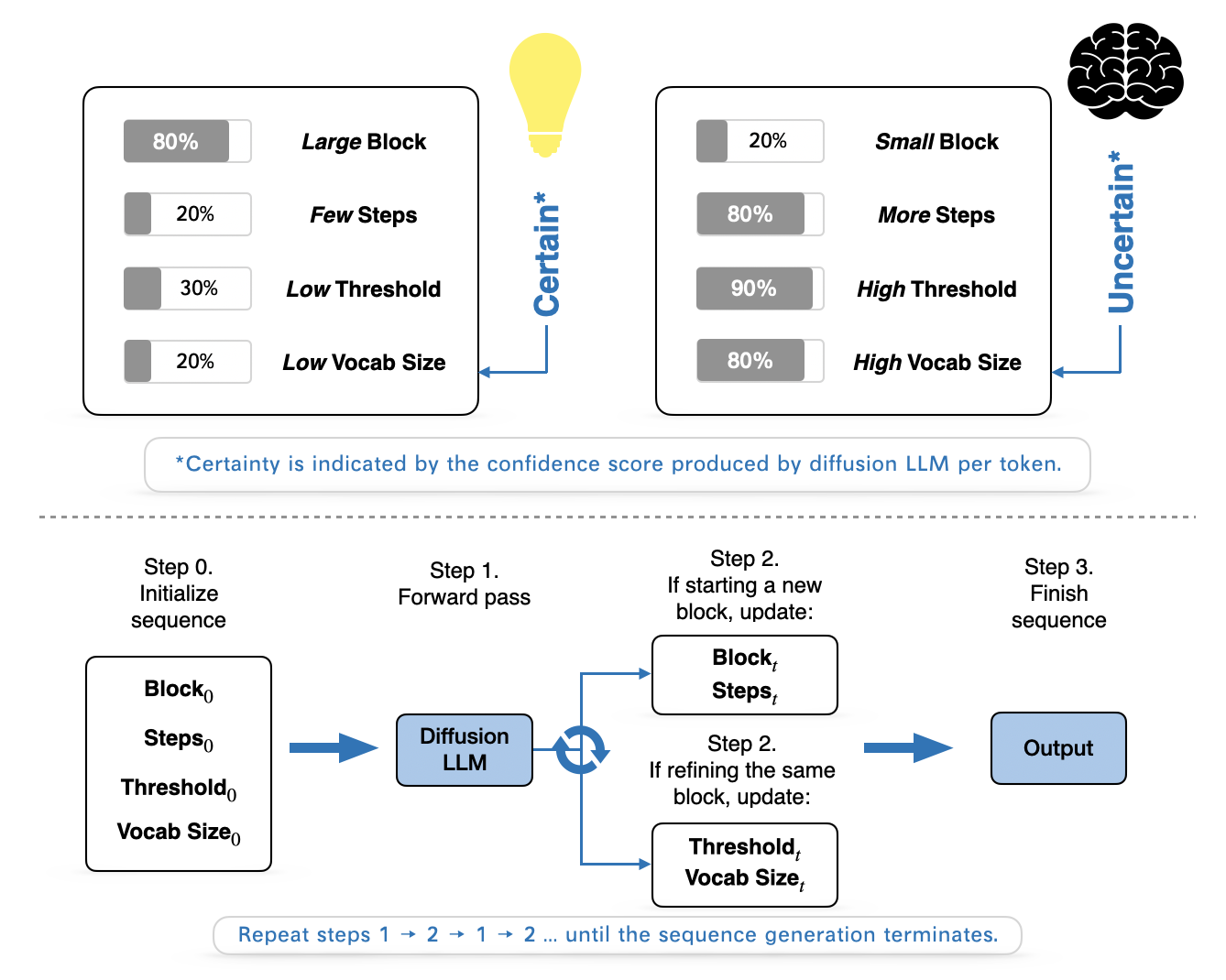}
  \vspace{-4mm}
  \caption{Overview of CadLLM’s adaptive controller. The controller dynamically updates various parameters based on a lightweight confidence and progress signals, replacing the static ones.}
  \label{fig:algo_overview}
  \vspace{-5mm}
\end{figure}

As illustrated in Figure~\ref{fig:algo_overview}, we utilize diffusion LLM's internal confidence for each token after each forward pass as the key signal to form a \textit{feedback loop}. The confidence score serves as a single shared signal, and four policies form a closed-loop controller over four complementary resource dimensions: (1) $B_t$: how many tokens to decode in parallel (scope), (2) $S_t$: how many refinement steps to allocate (depth), (3) $V_t$: how broad the candidate set is (breadth), and (4) $\tau_t$: when to commit and move on (aggressiveness). Together, they jointly regulate the compute-per-token from a single signal, allocating resources where uncertainty persists and saving them where predictions have stabilized.

\paragraph{Design Rationale.} All four policies use linear interpolation with clipping---a deliberate choice to isolate the benefit of \textit{adaptivity itself} from any particular schedule form. Linear interpolation is the simplest monotonic mapping from a bounded signal (confidence $\in [0,1]$) to a bounded resource budget, introducing no additional inference-time computation beyond a single multiply-add per policy per block. Our key contribution is demonstrating that even this minimal adaptivity substantially outperforms static schedules, establishing a strong empirical lower bound for adaptive control in dLLMs.

\subsection{Adaptive block sizes ($B_t$)}
Figure~\ref{fig:step_block_heatmap} shows that (i) confidence typically rises then plateaus within few refinements and (ii) unmasking difficulty varies across blocks. As demonstrated in the following Eq.~\ref{eq:block_adaptive}, we set $B_t$ proportional to the current average confidence $\bar c$: high‑confidence blocks expand in size to amortize computing cost of forward passes, whereas low‑confidence blocks shrink so the model can focus refinement where uncertainty concentrates. 
\begin{equation}
\small
B_t = \mathrm{clip}\!\big(B_{\min}+(B_{\max}-B_{\min})\cdot \bar c,~B_{\min},B_{\max}\big)   
\label{eq:block_adaptive}
\end{equation}
\noindent
Here $B_{\max}$ and $B_{\min}$ represents the maximum and minimum allowable blocks sizes, respectively. The mean confidence $\bar c$ is computed over a sliding window of the $\Delta$ most recent per-step confidence values. We use a short window ($\Delta{=}2$) for $B_t$ and $S_t$ so the controller reacts quickly to confidence shifts---critical because block and step decisions must adapt promptly to avoid premature commitment or wasted refinement. For $V_t$, a longer window ($\Delta{=}5$) smooths transient fluctuations, since vocabulary breadth is less sensitive to instantaneous confidence changes.

\subsection{Adaptive steps ($S_t$)}
We design the step size $S_t$ to be complementarily related to $\bar c$ and use it to pace the inner loop within the active block and also schedule the threshold $\tau_t$, as described in Eq.~\ref{eq:step_adaptive}. This allows low-confidence mask locations to trigger more refinement steps, while high-confidence locations to have fewer steps.
\begin{equation}
\small
S_t = \mathrm{clip}\!\big(S_{\text{base}}+(S_{\max}-S_{\text{base}})(1-\bar c),~S_{\text{base}},S_{\max}\big)
\label{eq:step_adaptive}
\end{equation}
\subsection{Adaptive vocabulary size ($V_t$)}
To amortize the cost of sampling from vocabulary,  we adapt a sampling size $V_t$ based on few key metric: phase, confidence, and token repetition. In specific, we use a larger vocabulary subset at the early generation stage or during the phase of uncertainty in token generation. We also allow larger vocab size when repetitive tokens generate ($r_t$). On the contrary we reduce its size for higher confidence situation to reduce compute cost. Eq.~\ref{eq:vocab_adaptive} defines the $V_t$ allocation mechanism.
\begin{equation}
\small
V_t = \mathrm{clip}\!\Big(V_{\text{phase}}(g_t)\cdot f_{\text{conf}}(\bar c)\cdot f_{\text{rep}}(r_t),~V_{\min},V_{\max}\Big)
\label{eq:vocab_adaptive}
\end{equation}
Here, $V_{\text{phase}}$ widens early and narrows late based on the generation progress (\% of tokens generated within a sequence) $g_t$; $f_{\text{conf}}$ expands under low confidence; and $f_{\text{rep}}$ widens briefly under repetitions. In practice, the vocabulary subset is selected by taking the top-$V_t$ token indices ranked by raw logit magnitude (before softmax); the softmax is then computed only over these $V_t$ logits, reducing cost from $O(|\mathcal{V}|)$ to $O(V_t)$ where $V_t$ is typically 35--1,000.
\paragraph{Same Token Repetition Detection.}
A naive controller may over-narrow $V_t$ once confidence is high, causing short loops (e.g., “the the the”). To avoid this, we introduce a lightweight \emph{repetition detector} that scans recent outputs and returns $r_t\in[0,1]$. When repetition spikes, $f_{\text{rep}}(r_t)>1$ briefly widens $V_t$ to restore diversity, then reverts as $r_t$ drops. This preserves fast decoding in stable regions while preventing local degeneracy.
\subsection{Adaptive threshold ($\tau_t$)}
Adaptive confidence threshold \textit{controls the token unmasking}. Early in a block or for low confidence, committing too many tokens in parallel by having a low threshold lead to poor quality output. Also, having a high threshold late in a block gates already stable tokens from being unmasked. As described in Eq.~\ref{eq:th_adaptive}, we define a progress‑aware threshold $\tau_t$ that is initially strict (when confidence is low) and relaxes eventually (as confidence rises and remains high), as more and more tokens are unmasked and uses it to gate unmasking within the selected $V_t$ candidates.
\vspace{-3mm}
\begin{equation}
\small
\tau_t = \tau_{\text{base}}(1-g_t)+\tau_{\min}g_t
\vspace{-3mm}
\label{eq:th_adaptive}
\end{equation}

\section{Experiments}
\label{sec:eval}

\subsection{Experimental Setup}

All experiments are run on a single NVIDIA H100 GPU. We evaluate on GSM8K (5-shot), MATH (4-shot), MBPP (3-shot, pass@1), and HumanEval (0-shot, pass@1) with generation lengths $g{=}256$ and $g{=}512$. We compare three decoding strategies: {CadLLM (ours)}, and fast-dLLM with factor and threshold based approaches. Further hyperparameter details are provided in the Appendix.

\newcommand{\besttps}[1]{\cellcolor{yellow!25}\textbf{#1}}
\newcommand{\besttime}[1]{\cellcolor{yellow!25}\textbf{#1}}
\newcommand{\bestnfe}[1]{\cellcolor{yellow!25}\textbf{#1}}
\newcommand{\bestacc}[1]{\cellcolor{green!18}\textbf{#1}}

\subsection{Analysis of Main Results}
Table~\ref{tab:main-results} reports throughput (tokens/s), accuracy, and latency across benchmarks on LLaDA. {CadLLM} achieves 1.1--2.28$\times$ higher throughput than Fast-dLLM baselines while maintaining accuracy on par with the strongest models, including on MATH. We also evaluate on DREAM across all four benchmarks with consistent 1.1--1.4$\times$ gains (see Appendix Table~\ref{tab:dream-full-results}). Adaptive control over blocks, steps, and vocabulary size reduces number of forward passes and compute cost without retraining. These gains persist at larger scales ($g=512$), demonstrating that {CadLLM} maintains both efficiency and accuracy across longer generations. 



\newcommand{\accspeed}[3]{\makecell{#1\% \\ \textcolor{blue}{#2}~(\textcolor{orange}{#3})}}

\newcommand{\tighttbl}{\setlength{\tabcolsep}{2.5pt}\renewcommand{\arraystretch}{0.95}}

\definecolor{bestbg}{RGB}{255,245,204} 
\newcommand{\bestcell}[1]{\cellcolor{bestbg}{#1}} 

\begin{table}[t]
\centering
\footnotesize
\tighttbl
\begin{tabularx}{\linewidth}{@{}>{\raggedright\arraybackslash}X c c c@{}}
\toprule
\textbf{Benchmark} &
\makecell{\textbf{CadLLM}\\\textbf{(ours)}} &
\makecell{\textbf{Fast-dLLM}\\\textbf{(factor)}} &
\makecell{\textbf{Fast-dLLM}\\\textbf{(threshold)}} \\
\midrule
\multirow{2}{*}{GSM8K}
& \bestcell{\accspeed{78.01}{120.07}{1.33$\times$}} & \accspeed{76.19}{119.18}{1.32$\times$} & \accspeed{79.00}{90.40}{1.00$\times$} \\
& \bestcell{\accspeed{75.44}{107.79}{1.37$\times$}} & \accspeed{74.45}{100.66}{1.27$\times$} & \accspeed{75.28}{78.77}{1.00$\times$} \\
\midrule
\multirow{2}{*}{MATH}
& \bestcell{\accspeed{32.06}{106.84}{1.34$\times$}} & \accspeed{32.22}{109.97}{1.38$\times$} & 
\accspeed{32.40}{79.58}{1.00$\times$} \\
& \bestcell{\accspeed{34.94}{117.21}{1.14$\times$}} & \accspeed{35.40}{111.86}{1.08$\times$} & \accspeed{32.06}{103.18}{1.00$\times$} \\
\midrule
\multirow{2}{*}{MBPP}
& \bestcell{\accspeed{24.00}{99.86}{1.37$\times$}} & \accspeed{21.20}{96.01}{1.31$\times$} & \accspeed{25.60}{73.15}{1.00$\times$} \\
& \bestcell{\accspeed{13.00}{104.62}{1.35$\times$}} & \accspeed{13.20}{100.73}{1.30$\times$} & \accspeed{14.20}{77.71}{1.00$\times$} \\
\midrule
\multirow{2}{*}{HEval}
& \bestcell{\accspeed{35.97}{220.81}{2.28$\times$}} & \accspeed{32.92}{132.28}{1.37$\times$} & \accspeed{37.19}{96.84}{1.00$\times$} \\
& \bestcell{\accspeed{43.29}{163.72}{1.74$\times$}} & \accspeed{41.46}{131.14}{1.38$\times$} & \accspeed{45.12}{94.41}{1.00$\times$} \\
\bottomrule
\end{tabularx}
\captionsetup{skip=4pt}
\caption{Comprehensive results on LLaDA-Instruct (single NVIDIA H100). For each benchmark, the \emph{upper row} is generation length 256 and the \emph{lower row} is 512. Shot settings: GSM8K (5-shot), MATH (4-shot), MBPP (3-shot, pass@1), HumanEval (0-shot, pass@1). Each cell shows accuracy (top, in \%) and throughput improvement (bottom: \textcolor{blue}{tokens/s} / \textcolor{orange}{improvement}). 
Improvement is relative to the Fast-dLLM threshold baseline at the same generation length.}
\vspace{-5mm}
\label{tab:main-results}
\end{table}



\subsection{Ablations and Analysis}
We conduct ablation studies to understand how each policy of CadLLM contribute to performance (Table~\ref{tab:ablations-gsm8k}). Additionally, we also investigate the effects of using different cache modes (Table~\ref{tab:cache-modes}), confidence calculation methods (Table~\ref{tab:appendix-softmax-vs-entropy}), as well as enabling an optional early stopping via Prophet method that is proposed in recent literature (Table~\ref{tab:appendix-prophet}).
We include the latter two in the Appendix. We also provide pairwise $B_t \times S_t$ ablations confirming their complementary roles (Appendix Table~\ref{tab:pairwise-ablation}) and a $\pm$20\% hyperparameter sensitivity sweep demonstrating stable operating regions (Appendix~\S\ref{subsec:hp-sensitivity}).

\sisetup{
  table-number-alignment = center,
  table-text-alignment = center,
  group-separator = {,},
  detect-weight = true,
  detect-inline-weight = math
}

\sisetup{
  group-separator = {,},
  detect-weight = true,
  detect-inline-weight = math
}

\begin{table}[h]
\centering
\footnotesize
\begin{tabular*}{\linewidth}{@{\extracolsep{\fill}} l
                               S[table-format=3.2, table-number-alignment=center]
                               S[table-format=4.2, table-number-alignment=center]
                               S[table-format=2.2, table-number-alignment=center, table-space-text-post = \%]
                               S[table-format=6.0, table-number-alignment=center]}
\toprule
\textbf{Mode} & {\textbf{Token/s} $\uparrow$} & {\textbf{Time (s)} $\downarrow$} & {\textbf{Accuracy} $\uparrow$} & {\textbf{NFE} $\downarrow$} \\
\midrule
ON         & 121.72 & 2501.10 & 78.01\% & 86816 \\
No $V_t$   & 119.67 & 2547.25 & 74.41\% & 86666 \\
No $S_t$   & 136.76 & 2235.68 & 76.12\% & 76955 \\
No $B_t$   & 111.19 & 2751.77 & 78.32\% & 92344 \\
No $\tau_t$&  34.57 & 8841.89 & 78.17\% & 337664 \\
OFF        &  34.32 & 8908.60 & 78.01\% & 337664 \\
\bottomrule
\end{tabular*}
\caption{GSM8K ablations at generation length 256. “ON” enables all four adaptive policies (adaptive block size, steps, vocabulary size, and threshold). Each subsequent row disables only the named policy. “OFF” disables all four adaptive policies.}
\vspace{-4mm}
\label{tab:ablations-gsm8k}
\end{table}

\paragraph{Adaptive Policy Ablations.} 
As shown in Table~\ref{tab:ablations-gsm8k}, disabling adaptive block size ($B_t$) reduces throughput by 8.6\%, while accuracy changes only slightly. This shows that block sizing mainly affects computation instead of accuracy. Larger blocks amortize computation cost when confidence is high, while smaller blocks localize refinement with some extra computation cost when confidence is low. 
Without adaptive steps ($S_t$), throughput increases by 12.3\% however, the accuracy drops by 1.9 percentage points. This indicates that steps alone balance inference speed and accuracy: fewer steps accelerate decoding with accuracy drop, and more steps preserve accuracy but incur higher computation cost.
Removing adaptive vocabulary size ($V_t$) lowers throughput by 1.7\% and sharply reduces accuracy by 4.6 percentage points, while NFE remains nearly unchanged. This confirms that adaptive vocabulary size acts as an exploration--exploitation lever: widening early or under uncertainty improves robustness, while narrowing later saves softmax cost without harming accuracy.  

Eliminating the progress-aware threshold($\tau_t$) collapses efficiency. Throughput falls by 71.6\%, latency increases by 254\%, and NFE jumps by 289\%, while accuracy remains nearly unchanged. This demonstrates that a dynamic commit gate is essential for practical runtime. The fully static \textbf{OFF} configuration shows nearly identical degradations, confirming that thresholding dominates the contribution to efficiency gains in the absence of other adaptive policies.  

\paragraph{Cache Modes (Prefix vs.\ Dual).}
Additionally, as shown in Table~\ref{tab:cache-modes}, our method is \emph{KV-cache agnostic}: it plugs into either prefix-only caching or dual (prefix+suffix) caching without any modification except for the few lines that replace the underlying cache methods.

\begin{table}[h]
\centering
\footnotesize
\begin{tabular*}{\linewidth}{@{\extracolsep{\fill}} l
                               S[table-format=3.2, table-number-alignment=center]
                               S[table-format=4.2, table-number-alignment=center]
                               S[table-format=2.2, table-number-alignment=center, table-space-text-post=\%]
                               S[table-format=1.0, table-number-alignment=center]}
\toprule
\textbf{Cache}  & \textbf{Token/s} $\uparrow$ & \textbf{Time (s)} $\downarrow$ & \textbf{Accuracy} $\uparrow$ & \textbf{NFE} $\downarrow$ \\
\midrule
\multirow{2}{*}{Prefix} & 145.20 & 2116.64 & 75.89\% & 82{,}677 \\
                        & 113.09 & 3155.55 & 76.72\% & 97{,}865 \\
\multirow{2}{*}{Dual}   & 121.72 & 2501.10 & 78.01\% & 86{,}816 \\
                        & 107.79 & 3261.45 & 75.44\% & 104{,}215 \\
\bottomrule
\end{tabular*}
\caption{GSM8K (5-shot). CadLLM with prefix-only and dual KV caching. For each cache mode, the \emph{upper row} is generation length 256 and the \emph{lower row} is 512.}
\vspace{-4mm}
\label{tab:cache-modes}
\end{table}

\section{Conclusions}
We present CadLLM, a \textit{training-free} adaptive decoding method for masked dLLMs that replaces fixed schedules with four confidence-driven policies: adaptive block size, step size, vocabulary size (with repetition guard), and commitment threshold. These policies focus compute on uncertain regions and skip stable ones while remaining model-agnostic and KV-cache compatible. Evaluations on multiple datasets and two model architectures (LLaDA, DREAM) demonstrate that CadLLM retains accuracy similar to SoTA while improving throughput by 1.1--2.28$\times$.


\section{Limitations}
\label{sec:limitations}
One limitation of CadLLM is that the accuracy and throughput improvements can potentially remain sensitive to a few hyperparameters that may often require careful tuning. Automated search of the calibration hyperparameters is an interesting future research. Further evaluations on multimodal generative tasks with dLLMs remains another interesting exploration direction. Additionally, all four policies in CadLLM deliberately use linear interpolation with clipping as the schedule form, chosen to isolate the benefit of adaptivity itself (see Design Rationale in \S\ref{sec:method}). While this minimal design already yields substantial gains, replacing it with learned policies (e.g., RL-based or Bayesian optimization over schedule forms) could further tighten the efficiency--quality trade-off.   
\bibliography{custom}

@misc{nie2025largelanguagediffusionmodels,
      title={Large Language Diffusion Models}, 
      author={Shen Nie and Fengqi Zhu and Zebin You and Xiaolu Zhang and Jingyang Ou and Jun Hu and Jun Zhou and Yankai Lin and Ji-Rong Wen and Chongxuan Li},
      year={2025},
      eprint={2502.09992},
      archivePrefix={arXiv},
      primaryClass={cs.CL},
      url={https://arxiv.org/abs/2502.09992}, 
}

@misc{ye2025dream7bdiffusionlarge,
      title={Dream 7B: Diffusion Large Language Models}, 
      author={Jiacheng Ye and Zhihui Xie and Lin Zheng and Jiahui Gao and Zirui Wu and Xin Jiang and Zhenguo Li and Lingpeng Kong},
      year={2025},
      eprint={2508.15487},
      archivePrefix={arXiv},
      primaryClass={cs.CL},
      url={https://arxiv.org/abs/2508.15487}, 
}

@article{ye2025dream,
  title={Dream 7b: Diffusion large language models},
  author={Ye, Jiacheng and Xie, Zhihui and Zheng, Lin and Gao, Jiahui and Wu, Zirui and Jiang, Xin and Li, Zhenguo and Kong, Lingpeng},
  journal={arXiv preprint arXiv:2508.15487},
  year={2025}
}

@misc{austin2021d3pm,
  title={Structured Denoising Diffusion Models in Discrete State-Spaces},
  author={Austin, Jacob and Johnson, Daniel D. and Ho, Jonathan and Tarlow, Daniel and van den Berg, Rianne},
  year={2021},
  eprint={2107.03006},
  archivePrefix={arXiv},
  primaryClass={cs.LG},
  url={https://arxiv.org/abs/2107.03006}
}

@misc{campbell2022ctmdm,
      title={A Continuous Time Framework for Discrete Denoising Models}, 
      author={Andrew Campbell and Joe Benton and Valentin De Bortoli and Tom Rainforth and George Deligiannidis and Arnaud Doucet},
      year={2022},
      eprint={2205.14987},
      archivePrefix={arXiv},
      primaryClass={stat.ML},
      url={https://arxiv.org/abs/2205.14987}, 
}

@misc{wu2025fastdllm,
  title={Fast-dLLM: Training-free Acceleration of Diffusion LLM by Enabling KV Cache and Parallel Decoding},
  author={Wu, Chengyue and Zhang, Hao and Xue, Shuchen and Liu, Zhijian and Diao, Shizhe and Zhu, Ligeng and Luo, Ping and Han, Song and Xie, Enze},
  year={2025},
  eprint={2505.22618},
  archivePrefix={arXiv},
  primaryClass={cs.CL},
  url={https://arxiv.org/abs/2505.22618}
}

@misc{li2025knowbefore,
  title        = {Diffusion Language Models Know the Answer Before Decoding},
  author       = {Li, Pengxiang and Zhou, Yefan and Muhtar, Dilxat and Yin, Lu and Yan, Shilin and Shen, Li and Liang, Yi and Vosoughi, Soroush and Liu, Shiwei},
  year         = {2025},
  eprint       = {2508.19982},
  archivePrefix= {arXiv},
  primaryClass = {cs.CL},
  doi          = {10.48550/arXiv.2508.19982},
  url          = {https://arxiv.org/abs/2508.19982}
}

@article{fan2018hierarchical,
  title={Hierarchical neural story generation},
  author={Fan, Angela and Lewis, Mike and Dauphin, Yann},
  journal={arXiv preprint arXiv:1805.04833},
  year={2018}
}

@misc{lu2026adablockdllmsemanticawarediffusionllm,
      title={AdaBlock-dLLM: Semantic-Aware Diffusion LLM Inference via Adaptive Block Size},
      author={Guanxi Lu and Hao Mark Chen and Yuto Karashima and Zhican Wang and Daichi Fujiki and Hongxiang Fan},
      year={2026},
      eprint={2509.26432},
      archivePrefix={arXiv},
      primaryClass={cs.LG},
      url={https://arxiv.org/abs/2509.26432},
}

@article{holtzman2018learning,
  title={Learning to write with cooperative discriminators},
  author={Holtzman, Ari and Buys, Jan and Forbes, Maxwell and Bosselut, Antoine and Golub, David and Choi, Yejin},
  journal={arXiv preprint arXiv:1805.06087},
  year={2018}
}

\section*{Appendix}
\label{sec:appendix}

\subsection{Algorithm}
Algorithm \ref{alg:sf} describes the detailed working principle of adaptive thresholding based token generation for CadLLM.

\begin{algorithm}[H]
\caption{CadLLM.\protect\footnotemark{}}
\label{alg:sf}
\begin{algorithmic}[1]
\Require model $p_\theta$, prompt $x_{\text{prompt}}$, length $g$
\Require $(B_{\min},B_{\max}), (S_{\min},S_{\max})$
\Require $(V_{\min},V_{\max})$, $(\tau_0,\tau_{\min})$
\Require window $\Delta$
\State $x \gets [x_{\text{prompt}};\texttt{[MASK]}^g]$
\State $pos \gets |x_{\text{prompt}}|$
\State $remain \gets g$
\State $C \gets [\,]$
\While{$remain>0$}
  \State $g_t \gets (g-remain)/g$
  \State $\bar c \gets \textsc{MeanConf}(C,\Delta)$
  \State $f_{\text{conf}}(\bar c)\in\{1.5,1.2,0.8,1.0\}$ by $\bar c$
  \State $r_t \gets$ repeat-score
  \State $f_{\text{rep}}(r_t)\gets1+2r_t$ if $r_t>0.5$, else $1$
  \State $B_t \gets \mathrm{clip}(B_{\min}+(B_{\max}-B_{\min})\bar c)$
  \State $S_t \gets \mathrm{clip}(S_{\min}+(S_{\max}-S_{\min})(1-\bar c))$
  \State $V_t \gets \mathrm{clip}(V_{\text{phase}}(g_t)\,f_{\text{conf}}(\bar c)\,f_{\text{rep}}(r_t))$
  \State $\tau_t \gets \tau_0(1-g_t)+\tau_{\min}g_t$
  \State $M \gets$ mask in $[pos:pos+B_t)$
  \For{$t'=1$ \textbf{to} $S_t$}
     \State $\ell \gets \textsc{Logits}(x)$
     \State $x_{\text{pred}} \gets \arg\max(\ell)$
     \State $p[i] \gets$ conf.\ of $x_{\text{pred}}[i]$ in $V_t$, $i\in M$
     \State commit $\gets \{i\in M:p[i]\ge\tau_t\}$
     \State $x[\text{commit}] \gets x_{\text{pred}}[\text{commit}]$
     \State append $\textsc{Mean}(p[i]|i\in M)$ to $C$
     \If{no \texttt{[MASK]} in $M$} \textbf{break} \EndIf
  \EndFor
  \State $pos \gets pos+B_t$
  \State $remain \gets remain-B_t$
\EndWhile
\State \Return $x$
\end{algorithmic}
\end{algorithm}

\footnotetext{For all initial variables and hyperparameters not specified here, see Table~\ref{tab:appendix-hp-ours-llada}.}

\subsection {Related Works}
\label{sec:related work}

\paragraph{Fast-dLLM.}
Fast-dLLM accelerates masked diffusion language models via confidence‑aware parallel decoding with (i) a fixed‑threshold rule that unmasks all positions with $c^i\!\ge\!\tau$ and (ii) a factor‑based rule selecting the largest $r$ with $(r{+}1)(1-c^{(r)})<\phi$, executed in fixed‑size blocks with a preset number of refinement steps \cite{wu2025fastdllm}. Our \textbf{CadLLM} departs from this static design by turning these knobs into policies: \emph{adaptive block size} (compute allocation to easy vs.\ hard blocks), \emph{adaptive steps} (precision–speed dial), \emph{adaptive vocabulary size} (confidence‑aware search breadth), and \emph{adaptive threshold} (phase-aware commit threshold). This training‑free controller closes the loop between signals and compute, improving throughput while preserving accuracy.

\paragraph{Early exiting in diffusion LMs.}
Concurrent work observes early answer convergence to propose \emph{Prophet}, that triggers an “all‑in’’ decode using the top‑2 confidence gap as a stop criterion \cite{li2025knowbefore}. Our focus is orthogonal: a training‑free compute allocation (adaptive blocks, steps, vocabulary size, and threshold). To further demonstrate the adaptability of CadLLM we integrate the early-exit strategy of prophet as an option for early block‑level generation. We leave Prophet disabled in main results because it overlaps with adaptive steps and reduces accuracy (see Appendix Sec.~\ref{subsec:add_result} for a detailed comparison).

\subsection {Hyperparameter Settings}

\begin{table}[H]
\centering
\footnotesize
\setlength{\tabcolsep}{0.75pt}
\begin{tabular}{l l}
\toprule
\textbf{Parameter} & \textbf{Value} \\
\midrule
Generation length $g$ & (256, 512) \\
Temperature $T$ & 0 \\
Remasking & low confidence \\
Confidence method & softmax \\
Confidence window ($B_t$, $S_t$) & $\Delta_{BS}$ = 2 \\
Confidence window ($V_t$) & $\Delta_{V}$ = 5 \\
Initial block size $B_0$ & 24 \\
Block size range $(B_{\min},B_{\max})$ & (4, 64) \\
Initial step $S_0$ & 24 \\
Step range $(S_{\text{min}}, S_{\max})$ & (24, 90) \\
Initial vocabulary size $V_0$ & 100 \\
Vocabulary size range $(V_{\text{min}}, V_{\max})$ & (35, 1000) \\
Initial threshold $\tau_0$ & 0.85 \\
Threshold minimum $\tau_{\text{min}}$ & 0.4 \\
Early stopping via Prophet & False \\
Repetition detector & window $w=8$ \\
Repetition detector & min length $=2$ \\
Caching & Dual Cache \\
\bottomrule
\end{tabular}
\caption{Hyperparameters for CadLLM used in Table~\ref{tab:main-results}.}
\label{tab:appendix-hp-ours-llada}
\end{table}
\begin{table}[H]
\centering
\small
\begin{tabular}{l l}
\toprule
\textbf{Parameter} & \textbf{Value} \\
\midrule
Generation length $g$ & (256, 512) \\
Temperature $T$ & 0 \\
Caching & DualCache \\
Selection rule & factor (f); threshold (t) \\
Factor / Threshold & $\phi=1.0$ (f); $\tau=0.9$ (t) \\
Confidence method & softmax \\
\bottomrule
\end{tabular}
\caption{Hyperparameters for Fast-dLLM baselines used in Table~\ref{tab:main-results}.}
\label{tab:hp-baselines}
\end{table}

\begin{table}[H]
\centering
\footnotesize
\setlength{\tabcolsep}{1pt}
\begin{tabular}{l l}
\toprule
\textbf{Parameter} & \textbf{Value} \\
\midrule
Generation length $g$ & (256, 512) \\
Temperature $T$ & 0 \\
Remasking & low confidence \\
Confidence method & softmax \\
Confidence window ($B_t$, $S_t$) & $\Delta_{BS}$ = 2 \\
Confidence window ($V_t$) & $\Delta_{V}$ = 5 \\
Initial block size $B_0$ & 24 \\
Block size range $(B_{\min},B_{\max})$ & (8, 48) \\
Initial step $S_0$ & 24 \\
Step range $(S_{\text{min}}, S_{\max})$ & (24, 70) \\
Initial vocabulary size $V_0$ & 100 \\
Vocabulary size range $(V_{\text{min}}, V_{\max})$ & (35, 1000) \\
Initial threshold $\tau_0$ & 0.85 \\
Threshold minimum $\tau_{\text{min}}$ & 0.4 \\
Early stopping via Prophet & False \\
Repetition detector & window $w=8$ \\
Repetition detector & min length $=2$ \\
Caching & Dual Cache \\
\bottomrule
\end{tabular}
\caption{Hyperparameter choices for CadLLM on DREAM.}
\label{tab:appendix-hp-ours-dream}
\end{table}

Tables~\ref{tab:appendix-hp-ours-llada} and~\ref{tab:hp-baselines} detail the hyperparameter settings for ours and the baseline Fast-dLLM, respectively. For both cases, we used the LLaDA-8B-Instruct model. Table~\ref{tab:appendix-hp-ours-dream} report the hyperparameter settings of our method used on DREAM-7B-Base model.

\subsection {Additional Experimental Analysis} \label{subsec:add_result}

\subsubsection {Confidence Method Choice}
We compare two confidence computations for our controller: maximum softmax probability (“Softmax”) and entropy-based confidence (“Entropy”). Results on GSM8K at $g{=}256$ (single H100) show that softmax is both faster and slightly more accurate (Table~\ref{tab:appendix-softmax-vs-entropy}), with substantially fewer function evaluations.

\begin{table}[H]
\centering
\footnotesize
\setlength{\tabcolsep}{4pt}
\begin{tabular*}{\linewidth}{@{\extracolsep{\fill}} l c c c c}
\toprule
\textbf{Mode} & \textbf{Token/s} $\uparrow$ & \textbf{Time(s)} $\downarrow$ & \textbf{Accuracy} $\uparrow$ & \textbf{NFE} $\downarrow$ \\
\midrule
Softmax  & 121.72 & 2501.10 & 78.01\% & 86{,}816 \\
Entropy  &  76.69 & 3981.49 & 77.03\% & 135{,}963 \\
\bottomrule
\end{tabular*}
\caption{Comparison of two confidence calculation modes (softmax vs. entropy) on GSM8K (5-shot), for generation length $g{=}256$.}
\label{tab:appendix-softmax-vs-entropy}
\end{table}

\subsubsection {With Early Commit}
We evaluate integrating \emph{Prophet} early-commit \cite{li2025knowbefore} into \textbf{CadLLM} (Table~\ref{tab:appendix-prophet}). On GSM8K at $g{=}256$ (single H100), enabling Prophet increases throughput and reduces function evaluations, but it also lowers accuracy, overlapping with the role of our adaptive-steps policy. As a result, we keep Prophet \emph{disabled by default}; it remains an optional speed knob for deployments that can trade a few accuracy points for lower latency. It is also noteworthy that the original idea of Prophet was explored on top of LLaDA. On the other hand, we are the first to exhaustively explore the validity of this idea in the context of fast-dLLM and added their orthogonal support to ours. In our evaluation, fast-dLLM remains the main baseline, as it is the current training-free  SoTA baseline with static decoding framework.

\begin{table}[H]
\centering
\footnotesize
\setlength{\tabcolsep}{4pt}
\begin{tabular*}{\linewidth}{@{\extracolsep{\fill}} l c c c c}
\toprule
\textbf{Mode} & \textbf{Token/s} $\uparrow$ & \textbf{Time(s)} $\downarrow$ & \textbf{Accuracy} $\uparrow$ & \textbf{NFE} $\downarrow$ \\
\midrule
Prophet  & 144.50 & 2113.72 & 75.28\% & 70{,}335 \\
Default & 121.72 & 2501.10 & 78.01\% & 86{,}816 \\
\bottomrule
\end{tabular*}
\caption{Comparison of enabling early stopping via Prophet vs. default mode on GSM8K (5-shot), $g{=}256$.}
\label{tab:appendix-prophet}
\end{table}


\subsubsection {Ablations with $B_t$, $S_t$, $V_t$}
Table~\ref{tab:ablation_test} demonstrates the contributions of each of the three components, namely, $B_t$, $S_t$, and $V_t$.

\begin{table}[H]
\centering
\footnotesize
\setlength{\tabcolsep}{0.6pt}
\begin{tabular*}{\linewidth}{@{\extracolsep{\fill}} l c c c c c}
\toprule
\textbf{Dataset} & \textbf{$B_t$} & \textbf{$S_t$} &\textbf{$V_t$} & \textbf{Token/s} $\uparrow$ & \textbf{Accuracy} $\uparrow$\\
\midrule
   & \xmark & \xmark & \xmark & 34.3 & 78.01 \\
   & \checkmark & \xmark & \xmark & 62.5 & 74.75 \\
GSM8K   & \xmark & \checkmark & \xmark & 34.8 & 78.01 \\
   & \xmark & \xmark & \checkmark & 33.7 & \textbf{78.17} \\
   & \checkmark & \checkmark & \checkmark & \textbf{122.9} & 78.01 \\
\bottomrule
\end{tabular*}
\caption{Ablation results on LLaDA with $g$ = 256.}
\label{tab:ablation_test}
\end{table}

\subsubsection {On Hyperparameter Sensitivity}
Diffusion based decoding output often depends on hyperparameter settings. However, their hyperparameter sensitivity is not unique to our method. For example, Fast-dLLM \cite{wu2025fastdllm} requires tuning to find the factor value, refresh interval, block length, and threshold settings. In practice, we found that the hyperparameters primarily depend on the model architecture rather than downstream task. Notably, we used one fixed hyperparameter set for all the task evaluations with LLaDA. Additionally, we now demonstrate the efficacy of a single fixed hyperparameter setting on a separate model, namely DREAM \cite{ye2025dream}, across all four benchmarks at both generation lengths in Table~\ref{tab:dream-full-results}. CadLLM consistently delivers throughput gains of 1.1--1.4$\times$ on DREAM, confirming that it can be used as an out-of-the-box plug-and-play solution without requiring per-task tuning.

\begin{table}[H]
\centering
\footnotesize
\tighttbl
\begin{tabular}{@{} l @{\hskip 12pt} c @{\hskip 12pt} c @{}}
\toprule
\textbf{Benchmark} &
\makecell{\textbf{CadLLM}\\\textbf{(ours)}} &
\makecell{\textbf{Fast-dLLM}\\\textbf{(threshold)}} \\
\midrule
\multirow{2}{*}{GSM8K}
& \bestcell{\accspeed{72.40}{86.14}{1.10$\times$}} & \accspeed{73.54}{78.24}{1.00$\times$} \\
& \bestcell{\accspeed{73.31}{103.86}{1.21$\times$}} & \accspeed{73.24}{85.77}{1.00$\times$} \\
\midrule
\multirow{2}{*}{MATH}
& \bestcell{\accspeed{35.98}{128.80}{1.28$\times$}} & \accspeed{37.36}{101.01}{1.00$\times$} \\
& \bestcell{\accspeed{37.20}{161.10}{1.20$\times$}} & \accspeed{38.38}{134.00}{1.00$\times$} \\
\midrule
\multirow{2}{*}{MBPP}
& \bestcell{\accspeed{52.20}{137.83}{1.40$\times$}} & \accspeed{53.40}{98.22}{1.00$\times$} \\
& \bestcell{\accspeed{53.00}{185.22}{1.23$\times$}} & \accspeed{54.80}{150.23}{1.00$\times$} \\
\midrule
\multirow{2}{*}{HEval}
& \bestcell{\accspeed{37.19}{104.31}{1.16$\times$}} & \accspeed{51.00}{89.91}{1.00$\times$} \\
& \bestcell{\accspeed{39.02}{102.66}{1.20$\times$}} & \accspeed{50.60}{85.62}{1.00$\times$} \\
\bottomrule
\end{tabular}
\captionsetup{skip=4pt}
\caption{Full results on DREAM-7B-Base (single NVIDIA H100) with a single fixed hyperparameter setting. For each benchmark, the \emph{upper row} is generation length 256 and the \emph{lower row} is 512. Each cell shows accuracy (top, in \%) and throughput improvement (bottom: \textcolor{blue}{tokens/s} / \textcolor{orange}{improvement}). Improvement is relative to the Fast-dLLM threshold baseline.}
\label{tab:dream-full-results}
\end{table}

\subsubsection {Scaling Behavior of Adaptive Vocab Size}
We examine how the adaptive vocabulary size $V_t$ behaves under substantially different tokenizer scales. Across both architectures evaluated, we find that the effective operating range of $V_t$ remains stable even when the total vocabulary size increases by more than $3\times$.

LLaDA uses a vocabulary of 50{,}257 tokens and achieves strong throughput--accuracy trade-offs with $V_t{=}1000$ (about 2\% of the vocabulary). DREAM, despite using a much larger vocabulary of 151{,}936 tokens, exhibits similarly stable behavior under the same $V_t{=}1000$ setting (about 0.65\%). This indicates that $V_t$ does not need to scale proportionally with vocabulary size.


\subsubsection{Pairwise $B_t \times S_t$ Ablation}
\label{subsec:pairwise-ablation}

We conduct the full pairwise $B_t \times S_t$ ablation (with $\tau_t$ and $V_t$ held ON throughout) on both GSM8K and HumanEval for LLaDA and DREAM. Recall that $B_t$ governs refinement \textit{scope} (how many tokens are revised per block), while $S_t$ governs refinement \textit{depth} (how many denoising steps are applied); together they represent orthogonal dimensions of the adaptive refinement mechanism.

\begin{table}[H]
\centering
\footnotesize
\setlength{\tabcolsep}{3pt}
\begin{tabular*}{\linewidth}{@{\extracolsep{\fill}} c c c c c c}
\toprule
\textbf{$B_t$} & \textbf{$S_t$} & \textbf{Token/s} $\uparrow$ & \textbf{Time(s)} $\downarrow$ & \textbf{Accuracy} $\uparrow$ & \textbf{NFE} $\downarrow$ \\
\midrule
\multicolumn{6}{c}{\textit{GSM8K ($g{=}256$, LLaDA)}} \\
\midrule
\checkmark & \checkmark & 110.21 & 2777.19 & 76.95 & 86{,}987 \\
\xmark & \checkmark & 98.52 & 3106.57 & 77.79 & 92{,}365 \\
\checkmark & \xmark & 127.33 & 2406.37 & 76.57 & 76{,}996 \\
\xmark & \xmark & 100.85 & 3035.45 & 77.03 & 89{,}542 \\
\midrule
\multicolumn{6}{c}{\textit{HumanEval ($g{=}256$, LLaDA)}} \\
\midrule
\checkmark & \checkmark & 206.07 & 311.32 & 32.93 & 9{,}228 \\
\xmark & \checkmark & 192.78 & 329.71 & 31.71 & 9{,}546 \\
\checkmark & \xmark & 213.19 & 300.06 & 31.10 & 8{,}373 \\
\xmark & \xmark & 186.98 & 338.72 & 29.88 & 9{,}203 \\
\midrule
\multicolumn{6}{c}{\textit{GSM8K ($g{=}256$, DREAM)}} \\
\midrule
\checkmark & \checkmark & 85.47 & 3934.71 & 71.65 & 183{,}695 \\
\xmark & \checkmark & 82.48 & 4077.22 & 72.55 & 187{,}157 \\
\checkmark & \xmark & 91.91 & 3659.20 & 70.05 & 167{,}716 \\
\xmark & \xmark & 30.89 & 10888.96 & 72.02 & 179{,}033 \\
\midrule
\multicolumn{6}{c}{\textit{HumanEval ($g{=}256$, DREAM)}} \\
\midrule
\checkmark & \checkmark & 104.88 & 399.13 & 35.37 & 20{,}298 \\
\xmark & \checkmark & 107.26 & 390.25 & 31.71 & 20{,}615 \\
\checkmark & \xmark & 113.89 & 367.53 & 36.59 & 18{,}949 \\
\xmark & \xmark & 33.48 & 1250.25 & 31.71 & 19{,}820 \\
\bottomrule
\end{tabular*}

\caption{Pairwise ablation between $B_t$ and $S_t$ (with $\tau_t$ and $V_t$ ON). The OFF/OFF configuration yields the lowest accuracy on HumanEval for both LLaDA (29.88\%) and DREAM (31.71\%), confirming $B_t$ and $S_t$ play complementary roles.}
\label{tab:pairwise-ablation}
\end{table}

\subsubsection{Hyperparameter Sensitivity Analysis ($\pm$20\%)}
\label{subsec:hp-sensitivity}

We present a systematic $\pm$20\% sensitivity sweep, varying one hyperparameter at a time while holding all others at their default values across both LLaDA and DREAM on GSM8K and HumanEval. Bold rows indicate the default setting. Experiments are run on 100 samples per task.

\paragraph{Default hyperparameter values.}

\begin{table}[H]
\centering
\footnotesize
\setlength{\tabcolsep}{2pt}
\begin{tabular*}{\linewidth}{@{\extracolsep{\fill}} l c c c c}
\toprule
& \textbf{LLaDA} & \textbf{LLaDA} & \textbf{DREAM} & \textbf{DREAM} \\
& \textbf{(GSM8K)} & \textbf{(HEval)} & \textbf{(GSM8K)} & \textbf{(HEval)} \\
\midrule
$B_{\text{init}}$ & 24 & 48 & 16 & 16 \\
$S_{\text{init}}$ & 24 & 16 & 24 & 24 \\
$B_{\max}$ & 64 & 96 & 48 & 48 \\
$S_{\max}$ & 90 & 32 & 70 & 70 \\
$B_{\min}$ & 4 & 12 & 8 & 8 \\
$\tau$ & 0.85 & 0.85 & 0.85 & 0.85 \\
\bottomrule
\end{tabular*}

\caption{Default hyperparameter values used in the sensitivity sweep.}
\label{tab:hp-defaults}
\end{table}

\begin{table}[H]
\centering
\footnotesize
\setlength{\tabcolsep}{2pt}
\begin{tabular*}{\linewidth}{@{\extracolsep{\fill}} c l c c c c}
\toprule
& \textbf{Variant} & \textbf{Token/s} $\uparrow$ & \textbf{Time(s)} $\downarrow$ & \textbf{Accuracy} $\uparrow$ & \textbf{NFE} $\downarrow$ \\
\midrule
\multicolumn{6}{c}{\textit{Sensitivity to $B_{\text{init}}$}} \\
19 & $-$20\% & 103.06 & 224.5 & 78.0 & 6{,}809 \\
\textbf{24} & \textbf{default} & \textbf{106.84} & \textbf{216.4} & \textbf{78.0} & \textbf{6{,}589} \\
29 & +20\% & 106.64 & 214.2 & 76.0 & 6{,}552 \\
\midrule
\multicolumn{6}{c}{\textit{Sensitivity to $S_{\text{init}}$}} \\
19 & $-$20\% & 109.07 & 211.4 & 80.0 & 6{,}568 \\
\textbf{24} & \textbf{default} & \textbf{107.53} & \textbf{215.0} & \textbf{78.0} & \textbf{6{,}589} \\
29 & +20\% & 108.42 & 213.6 & 76.0 & 6{,}602 \\
\midrule
\multicolumn{6}{c}{\textit{Sensitivity to $B_{\max}$}} \\
51 & $-$20\% & 104.22 & 221.6 & 80.0 & 6{,}671 \\
\textbf{64} & \textbf{default} & \textbf{108.09} & \textbf{213.9} & \textbf{78.0} & \textbf{6{,}589} \\
77 & +20\% & 107.48 & 216.3 & 75.0 & 6{,}681 \\
\midrule
\multicolumn{6}{c}{\textit{Sensitivity to $S_{\max}$}} \\
72 & $-$20\% & 106.11 & 217.2 & 80.0 & 6{,}583 \\
\textbf{90} & \textbf{default} & \textbf{108.82} & \textbf{212.5} & \textbf{78.0} & \textbf{6{,}589} \\
108 & +20\% & 105.29 & 219.7 & 75.0 & 6{,}629 \\
\midrule
\multicolumn{6}{c}{\textit{Sensitivity to $B_{\min}$}} \\
3 & $-$20\% & 106.67 & 216.8 & 79.0 & 6{,}592 \\
\textbf{4} & \textbf{default} & \textbf{106.29} & \textbf{217.5} & \textbf{78.0} & \textbf{6{,}589} \\
5 & +20\% & 108.79 & 212.5 & 77.0 & 6{,}539 \\
\midrule
\multicolumn{6}{c}{\textit{Sensitivity to $\tau$}} \\
0.68 & $-$20\% & 131.62 & 178.3 & 77.0 & 5{,}156 \\
\textbf{0.85} & \textbf{default} & \textbf{108.39} & \textbf{213.3} & \textbf{78.0} & \textbf{6{,}589} \\
1 & +20\% & 29.91 & 779.8 & 80.0 & 25{,}600 \\
\bottomrule
\end{tabular*}

\caption{Sensitivity analysis on LLaDA --- GSM8K ($g{=}256$, 100 samples).}
\label{tab:sens-llada-gsm8k}
\end{table}

\begin{table}[H]
\centering
\footnotesize
\setlength{\tabcolsep}{2pt}
\begin{tabular*}{\linewidth}{@{\extracolsep{\fill}} c l c c c c}
\toprule
& \textbf{Variant} & \textbf{Token/s} $\uparrow$ & \textbf{Time(s)} $\downarrow$ & \textbf{Accuracy} $\uparrow$ & \textbf{NFE} $\downarrow$ \\
\midrule
\multicolumn{6}{c}{\textit{Sensitivity to $B_{\text{init}}$}} \\
38 & $-$20\% & 223.44 & 164.2 & 45.0 & 5{,}484 \\
\textbf{48} & \textbf{default} & \textbf{236.14} & \textbf{157.5} & \textbf{45.0} & \textbf{5{,}264} \\
58 & +20\% & 245.23 & 150.7 & 42.0 & 5{,}126 \\
\midrule
\multicolumn{6}{c}{\textit{Sensitivity to $S_{\text{init}}$}} \\
13 & $-$20\% & 242.98 & 150.7 & 43.0 & 5{,}059 \\
\textbf{16} & \textbf{default} & \textbf{242.52} & \textbf{153.3} & \textbf{45.0} & \textbf{5{,}264} \\
19 & +20\% & 239.08 & 153.5 & 43.0 & 5{,}178 \\
\midrule
\multicolumn{6}{c}{\textit{Sensitivity to $B_{\max}$}} \\
77 & $-$20\% & 232.81 & 159.3 & 41.0 & 5{,}405 \\
\textbf{96} & \textbf{default} & \textbf{243.36} & \textbf{152.8} & \textbf{45.0} & \textbf{5{,}264} \\
115 & +20\% & 233.40 & 158.3 & 40.0 & 5{,}359 \\
\midrule
\multicolumn{6}{c}{\textit{Sensitivity to $S_{\max}$}} \\
26 & $-$20\% & 245.54 & 152.1 & 45.0 & 5{,}243 \\
\textbf{32} & \textbf{default} & \textbf{238.82} & \textbf{155.7} & \textbf{45.0} & \textbf{5{,}264} \\
38 & +20\% & 241.26 & 153.7 & 45.0 & 5{,}282 \\
\midrule
\multicolumn{6}{c}{\textit{Sensitivity to $B_{\min}$}} \\
10 & $-$20\% & 245.42 & 152.2 & 46.0 & 5{,}278 \\
\textbf{12} & \textbf{default} & \textbf{240.96} & \textbf{154.3} & \textbf{45.0} & \textbf{5{,}264} \\
14 & +20\% & 240.08 & 155.4 & 45.0 & 5{,}289 \\
\midrule
\multicolumn{6}{c}{\textit{Sensitivity to $\tau$}} \\
0.68 & $-$20\% & 256.61 & 143.7 & 42.0 & 5{,}061 \\
\textbf{0.85} & \textbf{default} & \textbf{241.72} & \textbf{153.8} & \textbf{45.0} & \textbf{5{,}264} \\
1 & +20\% & 50.26 & 730.2 & 47.0 & 25{,}600 \\
\bottomrule
\end{tabular*}

\caption{Sensitivity analysis on LLaDA --- HumanEval ($g{=}256$, 100 samples).}
\label{tab:sens-llada-heval}
\end{table}

\begin{table}[H]
\centering
\footnotesize
\setlength{\tabcolsep}{2pt}
\begin{tabular*}{\linewidth}{@{\extracolsep{\fill}} c l c c c c}
\toprule
& \textbf{Variant} & \textbf{Token/s} $\uparrow$ & \textbf{Time(s)} $\downarrow$ & \textbf{Accuracy} $\uparrow$ & \textbf{NFE} $\downarrow$ \\
\midrule
\multicolumn{6}{c}{\textit{Sensitivity to $B_{\text{init}}$}} \\
13 & $-$20\% & 84.04 & 303.3 & 69.0 & 14{,}315 \\
\textbf{16} & \textbf{default} & \textbf{82.64} & \textbf{308.5} & \textbf{69.0} & \textbf{14{,}315} \\
19 & +20\% & 84.81 & 300.6 & 69.0 & 14{,}315 \\
\midrule
\multicolumn{6}{c}{\textit{Sensitivity to $S_{\text{init}}$}} \\
19 & $-$20\% & 84.38 & 302.2 & 70.0 & 14{,}018 \\
\textbf{24} & \textbf{default} & \textbf{84.52} & \textbf{301.6} & \textbf{69.0} & \textbf{14{,}315} \\
29 & +20\% & 83.47 & 305.4 & 70.0 & 14{,}263 \\
\midrule
\multicolumn{6}{c}{\textit{Sensitivity to $B_{\max}$}} \\
38 & $-$20\% & 84.41 & 302.0 & 71.0 & 13{,}867 \\
\textbf{48} & \textbf{default} & \textbf{82.90} & \textbf{307.5} & \textbf{69.0} & \textbf{14{,}315} \\
58 & +20\% & 86.43 & 295.0 & 71.0 & 14{,}033 \\
\midrule
\multicolumn{6}{c}{\textit{Sensitivity to $S_{\max}$}} \\
56 & $-$20\% & 84.26 & 302.6 & 69.0 & 14{,}160 \\
\textbf{70} & \textbf{default} & \textbf{82.51} & \textbf{309.0} & \textbf{69.0} & \textbf{14{,}315} \\
84 & +20\% & 82.79 & 307.9 & 69.0 & 14{,}356 \\
\midrule
\multicolumn{6}{c}{\textit{Sensitivity to $B_{\min}$}} \\
6 & $-$20\% & 85.07 & 299.7 & 70.0 & 14{,}252 \\
\textbf{8} & \textbf{default} & \textbf{84.08} & \textbf{303.2} & \textbf{69.0} & \textbf{14{,}315} \\
10 & +20\% & 85.83 & 297.1 & 71.0 & 13{,}865 \\
\midrule
\multicolumn{6}{c}{\textit{Sensitivity to $\tau$}} \\
0.68 & $-$20\% & 93.01 & 274.1 & 71.0 & 12{,}811 \\
\textbf{0.85} & \textbf{default} & \textbf{83.87} & \textbf{303.9} & \textbf{69.0} & \textbf{14{,}315} \\
1 & +20\% & 48.32 & 527.6 & 76.0 & 25{,}600 \\
\bottomrule
\end{tabular*}

\caption{Sensitivity analysis on DREAM --- GSM8K ($g{=}256$, 100 samples).}
\label{tab:sens-dream-gsm8k}
\end{table}

\begin{table}[H]
\centering
\footnotesize
\setlength{\tabcolsep}{2pt}
\begin{tabular*}{\linewidth}{@{\extracolsep{\fill}} c l c c c c}
\toprule
& \textbf{Variant} & \textbf{Token/s} $\uparrow$ & \textbf{Time(s)} $\downarrow$ & \textbf{Accuracy} $\uparrow$ & \textbf{NFE} $\downarrow$ \\
\midrule
\multicolumn{6}{c}{\textit{Sensitivity to $B_{\text{init}}$}} \\
13 & $-$20\% & 112.38 & 227.2 & 41.0 & 12{,}032 \\
\textbf{16} & \textbf{default} & \textbf{110.20} & \textbf{231.7} & \textbf{41.0} & \textbf{12{,}032} \\
19 & +20\% & 110.86 & 230.3 & 41.0 & 12{,}032 \\
\midrule
\multicolumn{6}{c}{\textit{Sensitivity to $S_{\text{init}}$}} \\
19 & $-$20\% & 108.56 & 235.2 & 39.0 & 11{,}963 \\
\textbf{24} & \textbf{default} & \textbf{109.80} & \textbf{232.5} & \textbf{41.0} & \textbf{12{,}032} \\
29 & +20\% & 109.53 & 233.0 & 42.0 & 12{,}204 \\
\midrule
\multicolumn{6}{c}{\textit{Sensitivity to $B_{\max}$}} \\
38 & $-$20\% & 108.84 & 234.5 & 44.0 & 12{,}472 \\
\textbf{48} & \textbf{default} & \textbf{112.48} & \textbf{227.0} & \textbf{41.0} & \textbf{12{,}032} \\
58 & +20\% & 108.80 & 234.6 & 45.0 & 12{,}188 \\
\midrule
\multicolumn{6}{c}{\textit{Sensitivity to $S_{\max}$}} \\
56 & $-$20\% & 110.06 & 232.0 & 41.0 & 11{,}812 \\
\textbf{70} & \textbf{default} & \textbf{110.86} & \textbf{230.3} & \textbf{41.0} & \textbf{12{,}032} \\
84 & +20\% & 106.68 & 239.3 & 43.0 & 12{,}293 \\
\midrule
\multicolumn{6}{c}{\textit{Sensitivity to $B_{\min}$}} \\
6 & $-$20\% & 111.71 & 228.5 & 43.0 & 12{,}043 \\
\textbf{8} & \textbf{default} & \textbf{110.25} & \textbf{231.5} & \textbf{41.0} & \textbf{12{,}032} \\
10 & +20\% & 110.18 & 231.7 & 41.0 & 12{,}062 \\
\midrule
\multicolumn{6}{c}{\textit{Sensitivity to $\tau$}} \\
0.68 & $-$20\% & 122.46 & 208.4 & 43.0 & 10{,}873 \\
\textbf{0.85} & \textbf{default} & \textbf{112.23} & \textbf{227.5} & \textbf{41.0} & \textbf{12{,}032} \\
1 & +20\% & 51.03 & 500.3 & 43.0 & 25{,}600 \\
\bottomrule
\end{tabular*}

\caption{Sensitivity analysis on DREAM --- HumanEval ($g{=}256$, 100 samples).}
\label{tab:sens-dream-heval}
\end{table}

Across the five block and step hyperparameters ($B_{\text{init}}$, $S_{\text{init}}$, $B_{\max}$, $S_{\max}$, $B_{\min}$), accuracy varies by at most $\pm$3 pp on GSM8K and $\pm$5 pp on HumanEval within the $\pm$20\% range, and NFE shifts by less than 5\%, confirming that the default configuration lies in a stable operating region. The confidence threshold $\tau$ exhibits the expected monotonic trade-off: lowering it reduces NFE and speeds up inference, while raising it toward 1.0 forces near-exhaustive computation; the practical range $\tau \in [0.68, 0.90]$ remains well-behaved on both tasks. Crucially, \textbf{one fixed set of hyperparameters is used across all three of the four tasks for LLaDA (except HumanEval), and all four tasks for DREAM}, with no or very little per-task tuning.

\subsubsection{Statistical Uncertainty Analysis}
We have run all experiments on LLaDA with a total of \textit{three} different seeds to validate the stability of our results. We now report mean $\pm$ standard deviation for tokens/sec and accuracy in Table~\ref{tab:perf_stat_undertainty}. Note that (1) as NFE are the same across all runs for each task, we report the single NFE result without standard deviation; (2) since we derive accuracy for HumanEval (HEval) from the post-processing script, we do not obtain the standard deviation of accuracy and thus only report the single accuracy.

\begin{table}[!t]
\centering
\footnotesize
\setlength{\tabcolsep}{2.5pt}
\renewcommand{\arraystretch}{0.95}
\begin{tabular*}{\linewidth}{@{\extracolsep{\fill}} l c c c c}
\toprule
\textbf{Dataset} & \textbf{$g$} & \textbf{Token/s} & \textbf{NFE} &  \textbf{Accuracy} \\
\midrule
GSM8K & 256 & 120.07 {\tiny($\pm$0.51)} & 86816 & 77.79 {\tiny($\pm$1.14)} \\
 & 512 & 107.79 {\tiny($\pm$0.53)} & 103792 & 76.42 {\tiny($\pm$1.17)} \\
\midrule
MATH & 256 & 102.04 {\tiny($\pm$0.25)} & 423857 & 31.96 {\tiny($\pm$0.62)} \\
 & 512 & 117.21 {\tiny($\pm$0.34)} & 643229 & 34.42 {\tiny($\pm$0.63)} \\
\midrule
MBPP & 256 & 99.86 {\tiny($\pm$0.61)} & 27979 & 26.00 {\tiny($\pm$1.96)} \\
 & 512 & 104.62 {\tiny($\pm$0.43)} & 50776 & 12.8 {\tiny($\pm$1.49)} \\
\midrule
HEval & 256 & 220.81 {\tiny($\pm$1.62)} & 9066 & 31.09 {\tiny(NA)} \\
 & 512 & 163.72 {\tiny($\pm$1.64)} & 17527 & 37.20 {\tiny(NA)} \\
\bottomrule
\end{tabular*}

\caption{CadLLM: Throughput and accuracy statistical deviation analysis with multiple runs.}

\label{tab:perf_stat_undertainty}
\end{table}

\subsubsection {Failure Modes and Practical Guardrails}
Like other diffusion-based decoding methods, CadLLM exhibits failure modes under overly aggressive hyperparameter configurations. To characterize these limits, we perform 100-sample diagnostic runs on GSM8K ($g{=}256$), perturbing one hyperparameter at a time.
Under stressed settings, accuracy drops sharply on both models, as summarized in Table~\ref{tab:failure_modes}.

\begin{table}[H]
\centering
\footnotesize
\setlength{\tabcolsep}{2.5pt}
\renewcommand{\arraystretch}{0.95}
\begin{tabular*}{\linewidth}{@{\extracolsep{\fill}} l r r}
\toprule
\textbf{Setting} & \textbf{LLaDA (\%)} & \textbf{DREAM (\%)} \\
\midrule
$V_t{=}1$ & 8 & 1 \\
$\tau_0{=}0.3,\ \tau_{\min}{=}0.3$ & 35 & 28 \\
$S_0{=}6,\ S_{\max}{=}12$ & 29 & 20 \\
$B_0/B_{\max}/B_{\min}$ & 25 & 24 \\
\bottomrule
\end{tabular*}

\caption{Accuracy drops under hyperparameter stress on GSM8K ($g{=}256$).
Each row perturbs one hyperparameter.
For blocks,
$B_0/B_{\max}/B_{\min}
= \big(\text{LLaDA: }112/128/96;\ \text{DREAM: }80/96/64\big)$.}
\label{tab:failure_modes}
\end{table}

\noindent
These results show that extremely small vocabulary sizes, overly low thresholds, insufficient refinement steps, or excessively large blocks can all lead to severe accuracy collapse, consistent with premature commitment and under-refinement in dLLM.

To illustrate the qualitative failure pattern, below we include one representative output produced by LLaDA under an unstable configuration ($\tau_0{=}0.3,\ \tau_{\min}{=}0.3$):

\begin{tcolorbox}[
  colback=gray!5, colframe=gray!60, rounded corners,
  boxrule=0.5pt, left=6pt, right=6pt, top=4pt, bottom=4pt,
  width=\linewidth, fontupper=\footnotesize
]
\textbf{Question:}\smallskip\\
{\fontfamily{pcr}\selectfont
Kylar went to the store to buy glasses for his new apartment. One glass costs \$5, but every second glass costs only 60\% of the price. Kylar wants to buy 16 glasses. How much does he need to pay for them?}

\medskip
\textbf{Answer:}\smallskip\\
{\fontfamily{pcr}\selectfont
To determine the total cost for 16 glasses, we need to consider the cost of the first glass and the subsequent discounted.\par
\smallskip
-- First glass: \$5.00\\
-- Second glass: 0.60 * \$5.00 = \$3.00\\
-- Total:: \$ \$ .000 + \$ \$ .00 = \$ \$8.00\\
-- Total:: \$ \$8.00 + \$ \$3.00 = \$ \$11.00\\
-- Total: \$ \$ \$ \$ \$ .00 \$ \$ \$ \$ \$ \$ \$ \$ \$ \$ \$ \$}
\end{tcolorbox}




\noindent
These observations motivate simple practical guardrails: avoid overly restrictive $V_t$, maintain moderate initial thresholds, and allow sufficient refinement steps early in decoding. Within the empirically validated ranges reported in this appendix, CadLLM remains stable across architectures while consistently improving throughput.

\subsection{Qualitative Case Studies}

We compare answers produced by our controller in three settings: \emph{ON} (all four adaptive policies enabled), an \emph{ablated} variant with exactly one policy disabled, and \emph{OFF} (all four adaptive policies disabled). The answers are shown in Table~\ref{case: no vocab size}.

\begin{tcolorbox}[
  colback=gray!5, colframe=gray!60, rounded corners,
  boxrule=0.5pt, left=6pt, right=6pt, top=4pt, bottom=4pt,
  width=\linewidth, fontupper=\footnotesize
]
\textbf{Question:}\smallskip\\
{\fontfamily{pcr}\selectfont
Janet’s ducks lay 16 eggs per day. She eats three for breakfast every morning and bakes muffins for her friends every day with four. She sells the remainder at the farmers’ market daily for \$2 per fresh duck egg. How much in dollars does she make every day at the farmers’ market?}
\end{tcolorbox}


\newcommand{\casetablefmt}{%
  \renewcommand{\arraystretch}{1.25}%
  \setlength{\tabcolsep}{10pt}%
}

\begin{table}[h]
\casetablefmt
\centering
\footnotesize
\setlength{\tabcolsep}{2pt}
\begin{tabularx}{\linewidth}{@{}>{\bfseries}p{0.15\linewidth} >{\raggedright\arraybackslash}p{0.82\linewidth}@{}}
\toprule
ON &
To determine how much Janet makes every day at the farmers' market, we need to follow these steps:

1) Calculate the total number of eggs laid per day. 2) Subtract eggs eaten and baked. 3) Compute revenue.

Total eggs = 16. Breakfast = 3; baking = 4. Eggs left to sell = $16-3-4=9$. She sells each egg for \$2, so revenue $=9\times 2=18$. Therefore, Janet makes \$18 per day. \\
\addlinespace[6pt]
No $V_t$ &
To determine how much Janet makes every day at the farmers' market, we need to follow these steps:

1) Calculate the total number of eggs laid per day. 2) Subtract eggs eaten and baked. 3) Compute revenue.

Total eggs = 16. Breakfast = 3; baking = 4. Eggs left to sell = $16-3-4=9$. Price = \$2/egg. Revenue $=9\times 2=18$. Therefore, Janet makes \$18 per day. \\
\addlinespace[6pt]
OFF &
To determine how much Janet makes every day at the farmers' market, we need to follow these steps:

1) Total eggs laid per day: 16. 2) Eggs eaten for breakfast and baked for friends: $3+4=7$. 3) Eggs left to sell: $16-7=9$. 4) Revenue: $9 \times 2 = 18$.

Therefore, Janet makes \$18 every day. \\
\bottomrule
\end{tabularx}
\caption{Case study (adaptive vocabulary size ablation). Each variant is a single, full-width row.}
\label{case: no vocab size}
\end{table}

\begin{table}[h]
\casetablefmt
\centering
\footnotesize
\setlength{\tabcolsep}{2pt}
\begin{tabularx}{\linewidth}{@{}>{\bfseries}p{0.15\linewidth} >{\raggedright\arraybackslash}p{0.82\linewidth}@{}}
\toprule
ON &
To determine how much Janet makes every day at the farmers' market, we need to follow these steps:

1) Calculate the total number of eggs laid per day. 2) Subtract eggs eaten and baked. 3) Compute revenue.

Total eggs = 16. Breakfast = 3; baking = 4. Eggs left to sell = $16-3-4=9$. Price = \$2. Revenue $=9\times 2=18$. Therefore, \$18 per day. \\
\addlinespace[6pt]
No $S_t$ &
To determine how much Janet makes every day at the farmers' market, we need to follow these steps:

1) Calculate the total number of eggs laid per day. 2) Subtract eggs eaten and baked. 3) Compute revenue.

Total eggs = 16. Breakfast = 3; baking = 4. Eggs left for sale = $16-3-4=9$. Price = \$2/egg. Revenue $=9\times 2=18$. Therefore, \$18 per day. \\
\addlinespace[6pt]
OFF &
To determine how much Janet makes every day at the farmers' market, we need to follow these steps:

1) Total eggs laid per day: 16. 2) Eggs eaten for breakfast and baked for friends: $3+4=7$. 3) Eggs left to sell: $16-7=9$. 4) Revenue: $9\times \$2=\$18$.

Therefore, Janet makes \$18 every day. \\
\bottomrule
\end{tabularx}
\caption{Case study (adaptive steps ablation).}
\label{case: no steps}
\end{table}

\begin{table}[h]
\casetablefmt
\centering
\footnotesize
\setlength{\tabcolsep}{2pt}
\begin{tabularx}{\linewidth}{@{}>{\bfseries}p{0.15\linewidth} >{\raggedright\arraybackslash}p{0.82\linewidth}@{}}
\toprule
ON &
To determine how much Janet makes every day at the farmers' market, we need to follow these steps:

1) Calculate the total number of eggs laid per day. 2) Subtract eggs eaten and baked. 3) Compute revenue.

Total eggs = 16. Breakfast = 3; baking = 4. Eggs left to sell = $16-3-4=9$. Price = \$2. Revenue $=9\times 2=18$. Therefore, \$18 per day. \\
\addlinespace[6pt]
No $B_t$ &
To determine how much Janet makes every day at the farmers' market, we need to follow these steps:

1) Total eggs laid per day: 16. 2) Eggs eaten for breakfast and baked for friends: $3+4=7$. 3) Eggs left to sell: $16-7=9$. 4) Revenue: $9\times \$2=\$18$.

Therefore, Janet makes \$18 every day. \\
\addlinespace[6pt]
OFF &
To determine how much Janet makes every day at the farmers' market, we need to follow these steps:

1) Total eggs laid per day: 16. 2) Eggs eaten for breakfast and baked for friends: $3+4=7$. 3) Eggs left to sell: $16-7=9$. 4) Revenue: $9\times \$2=\$18$.

Therefore, Janet makes \$18 every day. \\
\bottomrule
\end{tabularx}
\caption{Case study (adaptive block size ablation).}
\label{case: no block}
\end{table}

\begin{table}[h!]
\casetablefmt
\centering
\footnotesize
\setlength{\tabcolsep}{2pt}
\begin{tabularx}{\linewidth}{@{}>{\bfseries}p{0.15\linewidth} >{\raggedright\arraybackslash}p{0.82\linewidth}@{}}
\toprule
ON &
To determine how much Janet makes every day at the farmers' market, we need to follow these steps:

1) Calculate the total number of eggs laid per day. 2) Subtract eggs eaten and baked. 3) Compute revenue.

Total eggs = 16. Breakfast = 3; baking = 4. Eggs left to sell = $16-3-4=9$. Price = \$2. Revenue $=9\times 2=18$. Therefore, \$18 per day. \\
\addlinespace[6pt]
No $\tau_t$ &
To determine how much Janet makes every day at the farmers' market, we need to follow these steps:

1) Total eggs laid per day: 16. 2) Eggs eaten for breakfast: 3. 3) Eggs baked for muffins: 4. 4) Eggs left to sell: $16-3-4=9$. 5) Revenue: $9\times \$2=\$18$.

Therefore, Janet makes \$18 at the farmers' market. \\
\addlinespace[6pt]
OFF &
To determine how much Janet makes every day at the farmers' market, we need to follow these steps:

1) Total eggs laid per day: 16. 2) Eggs eaten for breakfast and baked for friends: $3+4=7$. 3) Eggs left to sell: $16-7=9$. 4) Revenue: $9\times \$2=\$18$.

Therefore, Janet makes \$18 every day. \\
\bottomrule
\end{tabularx}
\caption{Case study (adaptive threshold ablation).}
\label{case: no threshold}
\end{table}

\paragraph{Adaptive vocabulary size (case study).}
In Table~\ref{case: no vocab size}, all three variants (ON, No adaptive vocabulary size, OFF) compute the correct answer ($18$). The qualitative difference is that \emph{ON} presents a concise, single-pass derivation with no redundant enumeration, whereas \emph{No adaptive vocabulary size} and \emph{OFF} tend to include extra scaffolding. This reflects the role of vocabulary size: keep breadth wide only when uncertainty is high, then narrow once scores concentrate. On harder prompts, this setting reduces token/phrase repetition by briefly widening the candidate set when the model hesitates, and then tightening it as confidence stabilizes.

\paragraph{Adaptive steps (case study).}
In Table~\ref{case: no steps}, all rows again reach $18$, but \emph{ON} keeps the reasoning succinct while \emph{No-adaptive steps} uses comparable logic with slightly more intermediate narration. Adaptive steps mainly protect quality on trickier questions: they allocate just enough refinement to ambiguous spans before committing, which avoids under-refinement errors (e.g., off‑by‑one arithmetic or prematurely truncated derivations) without forcing needless extra passes when the solution is already settled.

\paragraph{Adaptive block size (case study).}
Table~\ref{case: no block} shows that all settings deliver the same correct outcome, indicating the policy does not harm content fidelity on short arithmetic. Its qualitative value appears as sequences grow: \emph{ON} focuses work on the uncertain span only, preventing spillover from already-stable text and preserving step-by-step clarity. The ablated and all-off rows remain correct here, but on longer chains the adaptive span reduces cross-span interference and keeps explanations tighter.

\paragraph{Adaptive threshold (case study).}
In Table~\ref{case: no threshold}, each variant still returns $18$ with coherent steps. The adaptive threshold’s contribution is pacing: it raises the bar early (to avoid premature commitment when context is thin) and relaxes it late (to avoid redundant re-checking once the answer is evident). The \emph{No-adaptive threshold} row looks similar on this easy prompt, but on more ambiguous ones the adaptive gate prevents vacillation and unnecessary edits, yielding consistent derivations while preserving efficiency.

\end{document}